\documentclass[journal]{IEEEtran}
\usepackage{amsmath,amsfonts}
\usepackage{algorithmic}
\usepackage{algorithm}
\usepackage{array}
\usepackage[caption=false,font=normalsize,labelfont=sf,textfont=sf]{subfig}
\usepackage{textcomp}
\usepackage{stfloats}
\usepackage{url}
\usepackage{verbatim}
\usepackage{graphicx}
\usepackage{float}
\usepackage{subfig}
\usepackage{cite}
\usepackage{booktabs}
\usepackage{multirow}
\usepackage{makecell}
\usepackage{xcolor}
\usepackage{balance}
\usepackage{enumerate}
\begin{document}

\title{Structural Similarity in Deep Features: Image Quality Assessment Robust to Geometrically Disparate Reference}

\author{ Keke~Zhang,
	Weiling~Chen,~\IEEEmembership{Member,~IEEE},
	Tiesong~Zhao,~\IEEEmembership{Senior~Member,~IEEE}
	and~Zhou~Wang,~\IEEEmembership{Fellow,~IEEE}
	
	
	\thanks{This work was supported by the National Natural Science Foundation of China (62171134) and Natural Science Foundation of Fujian Province (2022J02015, 2022J05117). (\textit{Corresponding author: Tiesong Zhao.})}
	\thanks{Keke Zhang, Weiling Chen and Tiesong Zhao are with Fujian Key Lab for Intelligent Processing and Wireless Transmission of Media Information, College of Physics and Information Engineering, Fuzhou University, Fuzhou 350108, China and also with Fujian Science \& Technology Innovation Laboratory for Optoelectronic Information of China, Fuzhou 350108, China (e-mails: \{201110007, weiling.chen, t.zhao\}@fzu.edu.cn).}
	\thanks{Zhou Wang is with the Department of Electrical and Computer Engineering, University of Waterloo, Waterloo, ON N2L 3G1, Canada (e-mail: zhou.wang@uwaterloo.ca).}
}

\markboth{Journal of \LaTeX\ Class Files,~Vol.~14, No.~8, August~2021}%
{Shell \MakeLowercase{\textit{et al.}}: A Sample Article Using IEEEtran.cls for IEEE Journals}


\maketitle

\begin{abstract}
Image Quality Assessment (IQA) with references plays an important role in optimizing and evaluating computer vision tasks. Traditional methods assume that all pixels of the reference and test images are fully aligned. Such Aligned-Reference IQA (AR-IQA) approaches fail to address many real-world problems with various geometric deformations between the two images. Although significant effort has been made to attack Geometrically-Disparate-Reference IQA (GDR-IQA) problem, it has been addressed in a task-dependent fashion, for example, by dedicated designs for image super-resolution and retargeting, or by assuming the geometric distortions to be small that can be countered by translation-robust filters or by explicit image registrations. Here we rethink this problem and propose a unified, non-training-based Deep Structural Similarity (DeepSSIM) approach to address the above problems in a single framework, which assesses structural similarity of deep features in a simple but efficient way and uses an attention calibration strategy to alleviate attention deviation. The proposed method, without application-specific design, achieves state-of-the-art performance on AR-IQA datasets and meanwhile shows strong robustness to various GDR-IQA test cases. Interestingly, our test also shows the effectiveness of DeepSSIM as an optimization tool for training image super-resolution, enhancement and restoration, implying an even wider generalizability. \footnote{Source code will be made public after the review is completed.}
\end{abstract}


\section{Introduction}
\IEEEPARstart{I}{mage} Quality Assessment (IQA) aims at measuring visual quality of an image with or without an unimpaired or ideally perfect reference. This reference is generally indispensable in computer vision tasks to optimize and evaluate the qualities of their outputs. IQA community recognizes the above process as reference-based IQA. According to whether there exists geometric disparity between the reference and test images, we may categorize reference-based IQA into Aligned-Reference IQA (AR-IQA) and Geometrically-Disparate-Reference IQA (GDR-IQA). AR-IQA is designed for scenarios with registered, identically-sized reference and test images, with common applications such as image compression, transmission, and pixel-level processing. GDR-IQA includes IQA tasks for image super-resolution, retargeting and geometric transformation (shearing, rotation, translation, etc.). Both tasks play important roles in image processing systems \cite{QAsurvey}. However, there is currently no unified reference-based IQA index, resulting in limited application scenarios of related methods. Although large model can unify all reference-based IQA methods, it also fails to serve as an optimization tool (\emph{i.e.}, used in loss function) due to its high complexity.

\begin{figure}[t]
	\centering
	\includegraphics[width=8.8cm]{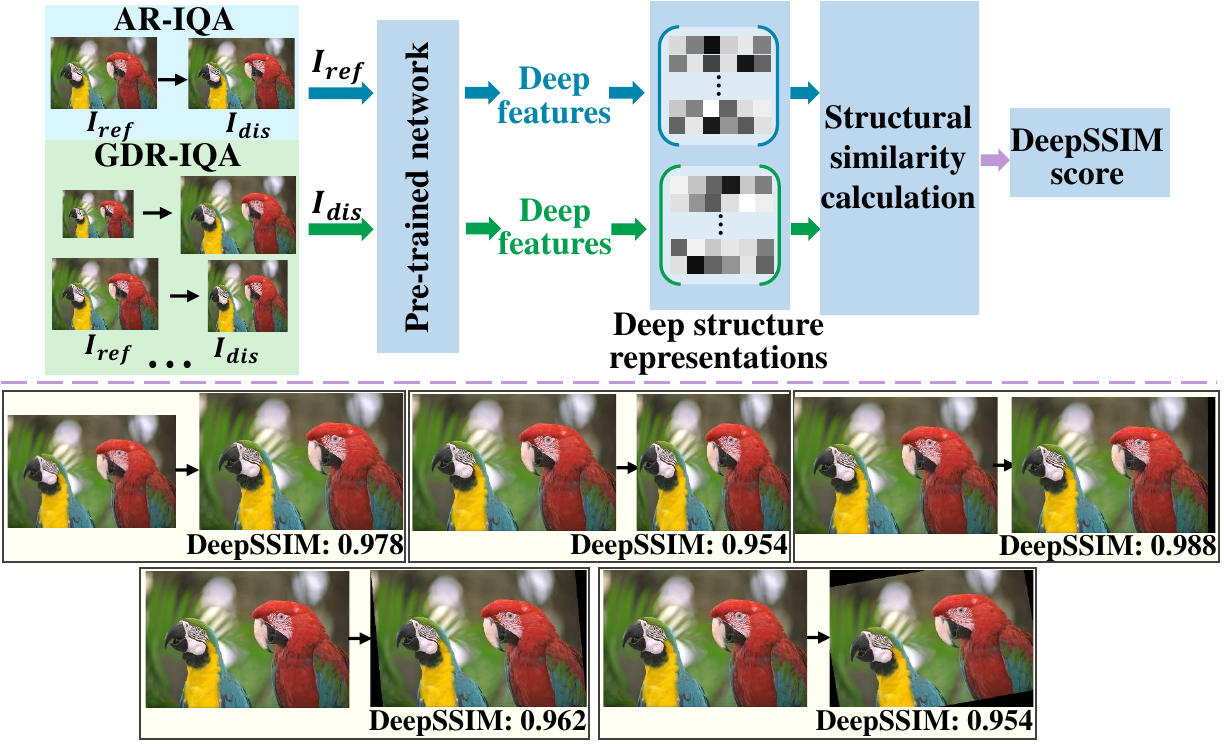}
	\caption{Our methodology and its performances when handling geometrically disparate references. We first construct deep structure representations based on deep features extracted by a pre-trained network. Then, we calculate the similarity between deep structure representations of reference and test images. The proposed DeepSSIM metric is robust to references with geometric disparities.}    
	\label{fig:basicidea}
\end{figure}

Over the past two decades, great success has been achieved on AR-IQA tasks. Representative methods include Structural Similarity (SSIM) \cite{SSIM}, Feature Similarity (FSIM) \cite{FSIM}, Learned Perceptual Image Patch Similarity (LPIPS) \cite{LPIPS} and so on. However, these methods do not consider image geometric deformation. Specifically, handcrafted-features-based AR-IQA methods are highly sensitive to geometric changes, even if the changes are tiny \cite{ma2018}. Hence, these AR-IQA methods do not handle well GDR-IQA tasks. In contrast, deep-learning-based AR-IQA methods show robustness to slight geometric deformation, but these methods cannot handle IQA tasks where the resolutions of reference and test images are inconsistent. To this end, various GDR-IQA methods have been proposed for specific tasks including Super-Resolution IQA (SRIQA) \cite{CNNSR,DeepSRQ,SFSN,SISAR,PSCT}, Image Retargeting Quality Assessment (IRQA) \cite{ARS,Ma2016,MLF,Jiang2020,INSEM} and Geometric Transformation Similarity Quality Assessment (GTSQA) (\emph{i.e.}, the similarity between source and transformed images) \cite{wang2005,cwssim,wang2010,ma2018}.

Rethinking the operation process and application scope of existing GDR-IQA methods, we find that they are task-dependent in similarity evaluation due to the resolution inconsistency problem between reference and test images. To be specific, SRIQA methods usually crop an image into patches to cope with the resolution inconsistency problem \cite{SISAR}. Similarly, IRQA methods generally adopt image registration preprocessing \cite{ARS,INSEM}. However, image cropping operations will lead to the loss of image information to some extent \cite{Leidaresize}. Moreover, registration estimations in IRQA are not exactly accurate \cite{MLF}. Besides, GTSQA methods are in their infancy and are only applicable to limited geometric transformation scenarios, \emph{e.g.}, up to four degrees of rotation \cite{cwssim} or scaling by a factor of 1.15 \cite{ma2018}. 

The scarcity of a task-independent reference-based IQA index limits the generalization of related methods, resulting in limited application scenarios. A unified index of reference-based IQA can address multiple tasks without loss of generalization ability. It also benefits the theoretical exploration of IQA problem and further development of computer vision tasks. To this end, we unify the essential problem of reference-based IQA task as: How to extract and compare the structure representations robust to references with geometric disparities? To the best of our knowledge, this problem has not yet been well addressed by IQA community.

As shown in Fig. \ref{fig:basicidea}, we propose a unified, non-training-based Deep Structural Similarity (DeepSSIM) metric that employs a deep structure representation of self-correlation between extracted deep features. Experiments reveal that the proposed representation is capable of characterizing image fidelity changes. Inspired by this, we calculate DeepSSIM by measuring the deep structural similarity between reference and test images. Our metric does not require further training; hence it does not involve the generalization problem.  

We summarize our main contributions as follows.
\begin{enumerate}[1)]
	\item We develop a deep structure representation by calculating the self-correlation between extracted deep features of an image. The similarity between deep structure representations is positively correlated with subjective scores in AR-IQA and shows robustness to geometric deformation in GDR-IQA.
	
	\item We propose a unified, non-training-based DeepSSIM metric, by calculating the similarity between deep structure representations of reference and test images. We also introduce the attention calibration strategy to alleviate attention deviation problem when extracting IQA-oriented deep features using a pre-trained network.

	\item Our DeepSSIM proves to achieve state-of-the-art evaluation performance on both AR-IQA and GDR-IQA tasks, and can be employed as an effective optimization tool for training computer vision tasks. To the best of our knowledge, it is the first metric that attempts to explore the essential issue in reference-based IQA, which brings new insights into general IQA tasks.
\end{enumerate}

The remainder of this paper is organized as follows. Section II introduces the related work. Section III elaborates the problem formulation, deep structure representation construction and implementation details to construct the proposed DeepSSIM metric. Section IV presents the experimental results. Section V concludes the paper.

\section{Related Work}

\subsection{AR-IQA Models}
Traditional AR-IQA models are designed to handle general distortions including Gaussian blur, JPEG compression and so on \cite{QAsurvey}. We empirically divide them into handcrafted-features-based and deep-learning-based models. 

\subsubsection{Handcrafted-Features-Based AR-IQA Models}
Wang \emph{et al.} \cite{SSIM} proposed the milestone IQA metric SSIM, which is based on the hypothesis that Human Visual System (HVS) is highly adapted to extract structural information from visual scenes. Given the great success of SSIM, several its variants have been proposed, such as multi-scale SSIM \cite{MSSSIM} and information content weighted SSIM \cite{IWSSIM}. With the in-depth research on HVS, image gradients are found to be beneficial for perceptual quality metric. In view of this, Zhang \emph{et al.} \cite{FSIM} proposed a feature similarity index based on phase congruency and gradient magnitude. Liu \emph{et al.} \cite{GSIM} designed an IQA metric that employs gradient similarity to measure the changes in image contrast and structure. Xue \emph{et al.} \cite{GMSD} developed an efficient IQA model that first computes gradient magnitude similarity maps of reference and distorted images and then calculates standard deviation between the two maps. 
\subsubsection{Deep-Learning-Based AR-IQA Models}
Inspired by the success of deep learning, researchers have developed massive deep-learning-based AR-IQA models. Kim \emph{et al.} \cite{KimIQA2017} proposed a Convolutional Neural Network (CNN)-based model, where the behavior of HVS is learned from the underlying data distribution of IQA datasets. Bosse \emph{et al.} \cite{BosseIQA2018} introduced a CNN-based framework, which can be used for full-reference and no-reference IQA with slight adaptions. Zhang \emph{et al.} \cite{LPIPS} proposed an LPIPS metric and demonstrated the effectiveness of deep features as a perceptual metric through massive experiments. Madhusudana \emph{et al.} \cite{CONTRIQUE} developed a deep IQA model using contrastive learning. Ding \emph{et al.} \cite{DISTS} proposed an IQA model that exhibits explicit tolerance to texture resampling by unifying structure and texture similarity. Xian \emph{et al.} \cite{SSHMPQA} designed an IQA model using structure separation and high-order moments in deep feature space. 

\subsection{GDR-IQA Models}
We review recent developments in GDR-IQA including SRIQA, IRQA and GTSQA models in this section. 

\subsubsection{SRIQA Models}
Existing reference-based SRIQA methods can also be divided into handcrafted-features-based and deep-learning-based methods. The former aims at capturing useful quality-aware features manually. Chen \emph{et al.} \cite{HYQM} proposed a hybrid quality metric for non-integer image interpolation, which extracts features from both reference-based and no-reference scenes. Zhou \emph{et al.} \cite{SFSN} designed a two-step metric to evaluate statistical naturalness and structural fidelity for image SR. Zhou and Wang \cite{SRIF} proposed a full-reference method to measure deterministic and statistical fidelity for image SR. Deep-learning-based methods have also been devoted to learning the mapping from images to quality scores. Zhao \emph{et al.} \cite{SISAR} designed a Deep Image SR Quality (DISQ) model by employing a two-stream CNN. The inputs of DISQ model are SR and Low-Resolution (LR) images, respectively. Zhang \emph{et al.} \cite{PSCT} developed a Perception-driven Similarity-Clarity Tradeoff (PSCT) model for SRIQA.
\subsubsection{IRQA Models}
Early IRQA methods calculate image similarity based on distances, such as bidirectional similarity \cite{BDS}, earth-mover's distance \cite{EMD} and SIFT flow \cite{SIFTflow}. However, Rubinstein \emph{et al.} \cite{RetargetMe} revealed that those distance-based methods do not correlate well with HVS. Liu \emph{et al.} \cite{CSim} developed an IRQA metric in a top-down manner, and designed a scale-space matching method to capture features. Fang \emph{et al.} \cite{IRSSIM} proposed an IRQA method based on structural similarity map of source and retargeted images. Hsu \emph{et al.} \cite{NRID} designed an IRQA metric that consider both perceptual geometric distortion and information loss. Zhang \emph{et al.} \cite{ARS} proposed a backward registration-based aspect ratio similarity metric for IRQA. Zhang \emph{et al.} \cite{MLF} designed a multiple-level feature-based quality measure for IRQA. Jiang \emph{et al.} \cite{Jiang2020} developed a deep-learning-based model for IRQA that integrates geometric and content features. Li \emph{et al.} \cite{INSEM} explored the impact of degraded instances on perceived quality and proposed an instance-semantics-based metric for IRQA. 

\subsubsection{GTSQA Models}
Limited work has been developed for geometric transformation similarity evaluation. Wang and Simoncelli \cite{wang2005} designed an adaptive linear system to decompose image distortions into a linear combination of components, which can handle tiny geometric distortions. Sampat \emph{et al.} \cite{cwssim} proposed a Complex Wavelet SSIM (CW-SSIM) index, which is robust to small geometric deformation, \emph{i.e.}, up to four degrees of rotations and seven pixels of translations. Nikvand and Wang \cite{wang2010} developed an image similarity index based on Kolmogorov complexity, which supports special geometric transformation such as global Fourier power spectrum scaling. Ma \emph{et al.} \cite{ma2018} constructed an end-to-end CNN that exhibits invariance to small geometric deformation, such as up to ten pixels of translation or up to five degrees of rotation.  

Despite recent advances, each of the above methods is generally designed for a single IQA task. There is lack of an investigation for common problem in reference-based IQA. In contrast, the proposed DeepSSIM metric can be applicable to the above IQA tasks simultaneously. 

\subsection{No-Reference IQA Models}
As a complement to the above AR-IQA and GDR-IQA models, we introduce no-reference traditional IQA, SRIQA and IRQA methods in this section.  

Early mainstream no-reference traditional IQA methods introduced Natural Scene Statistics (NSS), which represents statistical properties of natural scenes and might be altered in the presence of distortion. Moorthy and Bovik \cite{DIIVINE} utilized NSS features to propose a distortion identification-based image verity and integrity evaluation metric. Saad \emph{et al.} \cite{BLIINDS-II} developed an IQA algorithm using an NSS model of discrete cosine transform coefficients. Recently, deep-learning-based no-reference traditional IQA methods have emerged. Ying \emph{et al.} \cite{PaQ-2-PiQ} designed a deep region-based architecture that learns to produce global and local qualities. Chen \emph{et al.} \cite{FPR} introduced an IQA method by hallucinating feature-level reference information. Zhang \emph{et al.} \cite{weixia2023} built continual learning for IQA, where a model learns continually from a stream of IQA datasets. Liu \emph{et al.} \cite{BIQAMD} presented a deep blind IQA method based on multi-scale spatial filtering.

No-reference SRIQA methods focus on measuring qualities of SR images without considering reference information. Fang \emph{et al.} \cite{CNNSR} proposed a blind quality evaluation by an 8-layer CNN. Zhou \emph{et al.} \cite{DeepSRQ} designed a no-reference IQA using a two-stream CNN, which contains two subcomponents for processing structure and texture SR images, respectively. 
However, there are few no-reference IRQA methods given that distortions of IRQA tasks are highly correlated to reference images, such as geometric distortion and information loss. Ma \emph{et al.} \cite{Ma2016} proposed a no-reference method for retargeted images based on pairwise rank learning approach.  

Although no-reference IQA methods are readily utilized to IQA tasks, their performance is usually inferior to reference-based IQA methods and may not be fully reliable \cite{QAsurvey,DegradedReference}.

\section{Proposed DeepSSIM Metric}
\subsection{Problem Formulation}
AR-IQA comprises handcrafted-features-based and deep-learning-based methods. The former depends on the geometric consistency between reference and test images. In contrast, deep-learning-based AR-IQA methods show robustness to slight geometric deformation (as shown in Section IV), but still relies on the resolution consistency between two images. The above process can be expressed as:
\begin{equation}
\begin{aligned}
Q_{\rm AR}=M_{\rm AR}(X,Y),\\
s.t.\ g(X)=g(Y)\ or \ s(X)=s(Y),
\end{aligned}
\label{func:AR-IQA}
\end{equation}  
where $X$ and $Y$ represent an unimpaired reference and a test image, respectively. $M_{\rm AR}$ denotes an AR-IQA model to predict $Q_{\rm AR}$. $s(\cdot)$ means the size of an image. $g(\cdot)$ implies image geometric properties including the contour, location and other quantitative attributes of objects on the image. Accordingly, the requirement for geometric consistency is higher than that for size consistency.

As a contrast, GDR-IQA methods can be described as:
\begin{equation}
\begin{aligned}
Q_{\rm GDR}=M_{\rm GDR}(X',Y),
\forall X',Y,\ g(X')\neq g(Y),
\label{func:GDR-IQA}
\end{aligned}
\end{equation}  
where $X'$ refers to an LR image for image SR or a source image for image retargeting or geometric transformation. Obviously, in image SR, retargeting and shearing transformation processes, the resolution of $Y$ is changed compared with that of $X'$. In image rotation and translation transformations, the geometric properties of $X'$ and $Y$ are inconsistent even though their resolutions are identical. $M_{\rm GDR}$ denotes a GDR-IQA model, which can be $M\_SR$ for SRIQA, $M\_R$ for IRQA and $M\_T$ for GTSQA. $Q_{\rm GDR}$ is the corresponding quality score. Summarizing existing GDR-IQA methods, they can be separately represented as:
\begin{equation}
Q_{SR}=M\_SR(X',Y)=M\_SR(P(X'),P(Y)),
\label{func:SRIQA}
\end{equation} 
where $P(\cdot)$ denotes crop an image into patches operation;  
\begin{equation}
Q_R = M\_R(X',Y)=M\_R({\rm Regis}(X',Y)),
\label{func:IRQA}
\end{equation}
where ${\rm Regis}$ implies image registration preprocessing, such as backward registration, SIFT flow and so on;
\begin{equation}
Q_T = M\_T(X',Y), s.t.\ Y\gets T(X'),
\label{func:geometric}
\end{equation}
where $T(X')$ represents special transformation, such as up to six degrees of rotation.

Obviously, existing GDR-IQA methods are task-dependent and differ in task-oriented image preprocessing operations or special constraints for the distorted images. They lack the exploration for a general IQA problem: Is it possible to evaluate the structural similarity between artifact-free reference and test images, regardless of the existence of geometry consistency? To this end, we are dedicated to developing a unified metric in Eq. (\ref{func:our1}):
\begin{equation}
Q=M(\mathbb{X},Y),\ \mathbb{X}\in\{X,X'\},
\label{func:our1}
\end{equation} 
where $\mathbb{X}$ denotes an artifact-free reference image with arbitrary resolution compared to a distorted image $Y$. $M$ represents a unified IQA model. 

Inspired by SSIM, we leverage deep structural similarity to develop a unified IQA metric and name it as DeepSSIM. Denote the deep structure representation of an image $I$ as $R_{\rm DS}(I)$, Eq. (\ref{func:our1}) can be modified as: 
\begin{equation}
Q_{\rm DeepSSIM}=M_{\rm DeepSSIM}(R_{\rm DS}(\mathbb{X}),R_{\rm DS}(Y)).
\label{func:our2}
\end{equation}

\subsection{Deep Structure Representation Construction}
In this subsection, we construct an effective representation of $R_{\rm DS}(I)$ that fulfills the above four conditions. A straightforward idea to construct $R_{\rm DS}(I)$ is to find a deep mapping of structural information in SSIM. However, SSIM compares structure information in pairwise windows that are unfortunately inaccessible in GDR-IQA tasks. Moreover, considering that there are various geometric deformations in GDR-IQA, $R_{\rm DS}(I)$ cannot be constructed directly in spatial or frequency domain. Based on the above considerations, we construct $R_{\rm DS}(I)$ in feature domain.

Obviously, the feature extractor for $R_{\rm DS}(I)$ should be able to extract comprehensive perceptual information and exhibit invariance to geometric deformation to some extent. It should also be compatible with arbitrary size of images for GDR-IQA. In view of the above constraints, Transformer networks represented by ViT \cite{ViT} and Swin Transformer \cite{Swin} cannot be used here. In this paper, we employ a pre-trained CNN that satisfies the above requirements. First, CNN contains intrinsic invariance to geometric deformation \cite{CNNinvariance}. Second, CNN allows feature extraction using images with arbitrary size given its weight sharing and local connectivity mechanisms. Third, the effectiveness of using deep features from a pre-trained CNN to design perceptual metric has been demonstrated by \cite{LPIPS}.

There are various pre-trained CNNs including AlexNet \cite{AlexNet}, VGG16 \cite{VGG}, VGG19 \cite{VGG}, ResNet50 \cite{ResNet}, ResNet101 \cite{ResNet}, etc. We conduct massive experiments on eleven benchmark IQA datasets and find that employing the pre-trained VGG16 as our feature extractor usually achieves the optimal performance. This phenomenon is also consistent with the fact that IQA community usually employ the pre-trained VGG16 network as a backbone \cite{LPIPS, DISTS, weixia2023}. Consequently, we leverage the pre-trained VGG16 as the feature extractor and formulate the feature representation $F(I)$ with features extracted from the first convolutional layer of the last stage of VGG16 (labeled \emph{conv5\_1}):
\begin{equation}
F(I)=\{m_c^{(5)};c=1,...,512\},
\label{func:F(I)}
\end{equation}
where $m_c^{(5)}$ implies the feature maps extracted from \emph{conv5\_1}. The reason of using \emph{conv5\_1} here is that feature statistics from previous layers seem to capture basic intensity and color information, and those from posterior layers summarize the shape and structure information \cite{DISTS,textureGatys}. 

We cannot directly employ $F(I)$ as $R_{\rm DS}(I)$, since the size of $F(I)$ depends on the size of its corresponding image, which does not meet \emph{Condition 2} and poses difficulties to the similarity calculation of GDR-IQA tasks. Thus, we get back to HVS properties that reflect image structures. \cite{wangbook} reveals that samples taken from image signals have strong dependencies amongst themselves, and these dependencies carry important information about object structures. Inspired by this fact, we represent image structure by the self-correlation of its deep features in IQA. It is calculated as a Gram matrix:
\begin{equation}
R_{\rm DS}(I)=F(I)\cdot F(I)^T.
\label{func:DS}
\end{equation}
Although Gram matrix using deep features has been used for texture synthesis and style transfer \cite{textureGatys,styleGatys}, we are the first to validate its effectiveness in IQA tasks.

\subsection{The Proposed DeepSSIM}
\begin{figure*}[t]
	\centering
	\includegraphics[width=17cm]{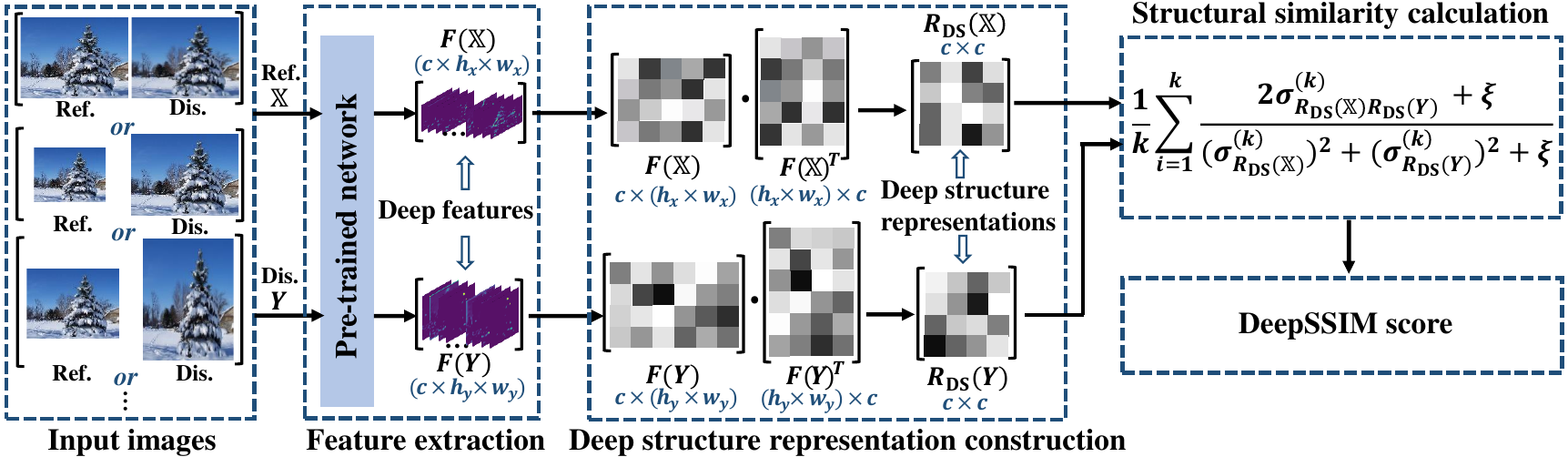}
	\vspace{-2mm}
	\caption{Our proposed DeepSSIM metric. First, we extract deep features ($F(\mathbb{X}), F(Y)$) of reference and distorted images ($\mathbb{X},Y$) from a pre-trained network. Second, we construct deep structure representations ($R_{\rm DS}(\mathbb{X}), R_{\rm DS}(Y)$) based on $F(\mathbb{X})$ and $F(Y)$. Third, we compute the structural similarity between reference and distorted images by comparing their deep structure representations ($R_{\rm DS}(\mathbb{X}), R_{\rm DS}(Y)$).}  
	\label{fig:framework}
\end{figure*}

With deep feature representations, we propose the overall framework of DeepSSIM as Fig. \ref{fig:framework}. First, it extracts deep features $F(\mathbb{X})$ and $F(Y)$ from a pre-trained network (VGG16 in this work). Second, it constructs deep structure representations $R_{\rm DS}(\mathbb{X})$ and $R_{\rm DS}(Y)$ based on the extracted features. Third, it calculates the structural similarity between images by comparing their deep structure representations. 

Inspired by the form of SSIM, we employ the correlation between deep structure representations of a distorted image and its reference image to construct our DeepSSIM metric. Aimed at a high accuracy, we leverage local correlations between $R_{\rm DS}(\mathbb{X})$ and $R_{\rm DS}(Y)$ and then average the resulting matrix as the finial DeepSSIM score. That is:
\begin{equation}
Q_{\rm DeepSSIM}=\frac{1}{k} \sum_{i=1}^{k}\frac{2\sigma_{R_{\rm DS}(\mathbb{X})R_{\rm DS}(Y)}^{(k)}+\xi}{(\sigma_{R_{\rm DS}(\mathbb{X})}^{(k)})^2+(\sigma_{R_{\rm DS}(Y)}^{(k)})^2+\xi},
\label{func:DeepSSIM}
\end{equation}
where $\sigma_{R_{\rm DS}(\mathbb{X})R_{\rm DS}(Y)}^{(k)}$ denotes the covariance between ${R_{\rm DS}(\mathbb{X})}^{(k)}$ and ${R_{\rm DS}(Y)}^{(k)}$. $(\sigma_{R_{\rm DS}(\mathbb{X})}^{(k)})^2$ and $(\sigma_{R_{\rm DS}(Y)}^{(k)})^2$ represent the variance of ${R_{\rm DS}(\mathbb{X})}^{(k)}$ and ${R_{\rm DS}(Y)}^{(k)}$, respectively. $\xi$ is a small constant to prevent zero denominator. In practice, the local correlations are calculated within 4$\times$4 windows in feature domain rather than spatial domain. We regard this version as the standard version of the DeepSSIM metric.
\subsubsection{Introducing DeepSSIM-Lite}
Considering the efficiency of the proposed DeepSSIM metric as an optimization tool for training computer vision tasks, we present its lightweight version and name it DeepSSIM-Lite (DSL). Specifically, we remove the attention calibration strategy and change the correlation calculation method between $R_{\rm DS}(\mathbb{X})$ and $R_{\rm DS}(Y)$ from local to global, \emph{i.e.}, set $k$ in Eq. (\ref{func:DeepSSIM}) to 1.   

\begin{table}[t]
	\centering
	\setlength{\abovecaptionskip}{0cm}
	\setlength{\belowcaptionskip}{-0.2cm}
	\setlength{\tabcolsep}{1mm}{
		\caption{Details of the utilized datasets}
		\label{table:datasets}
		\begin{tabular}{@{}c|c|c|c|c@{}}
			\toprule
			Task category&Datasets&\makecell{\#Reference\\ images}&\makecell{\#Distorted\\ images}&\makecell{Judgement\\ type}   \\ 
			\midrule
			&LIVE &29  &779 & DMOS\\ 
			&CSIQ&30&866&DMOS \\
			&TID2013&25&3000&MOS\\
			\multirow{-4}{*}{AR-IQA}&KADID-10k&81&10125&DMOS\\
			\midrule
			&CVIU&30&1620&MOS\\
			&QADS&20&980&MOS \\
			&SISAR&100&12600&MOS\\
			\cmidrule(l){2-5}
			&CUHK&57&171&MOS\\
			&RetargerMe&37&296&\#votes\\
			\multirow{-7}{*}{GDR-IQA}&NRID&35&175&\#votes \\
			\midrule
			\makecell{AR-IQA and\\ GDR-IQA}&\makecell{PIPAL\\ (Training set)}&200&23200&MOS\\
			
			\bottomrule
	\end{tabular}}
\end{table}  

\begin{table*}[ht]
	\centering
	\setlength{\abovecaptionskip}{0cm}
	\setlength{\belowcaptionskip}{-0.2cm}
	\setlength{\tabcolsep}{1.5mm}{

		\caption{Performance comparison on AR-IQA datasets and PIPAL dataset. We mark the training-based methods in italics and present their generalization performance. The top two results are labeled in bold and underline}
		\label{table:traditional}
		\begin{tabular}{@{}c|c|cc|cc|cc|cc||cc@{}}
			\toprule
			&&\multicolumn{2}{c|}{LIVE}&\multicolumn{2}{c|}{CSIQ}&\multicolumn{2}{c|}{TID2013}&\multicolumn{2}{c||}{KADID-10k}&\multicolumn{2}{c}{PIPAL} \\ 
			\cmidrule(l){3-12} 
			\multirow{-2}{*}{Algorithms}&\multirow{-2}{*}{\makecell{Support\\any size}}&PLCC&SRCC&PLCC&SRCC&PLCC&SRCC&PLCC&SRCC&PLCC&SRCC\\ 
			\midrule
			SSIM&$\times$&0.921&0.923&0.856&0.870&0.746&0.721&0.751&0.754&0.529&0.518\\
			\midrule
			FSIM&$\times$&\textbf{0.960}&\underline{0.963}&0.903&0.917&0.859&0.802&0.829&0.832&0.638&0.605\\
			\midrule
			GMSD&$\times$&\textbf{0.960}&0.960&\underline{0.947}&\textbf{0.950}&0.859&0.804&0.843&0.843&0.656&0.609\\
			\midrule
			\emph{LPIPS}&$\times$&0.939&0.940&0.899&0.881&0.471&0.394&0.746&0.738&0.654&0.613\\
			\midrule
			\emph{PaQ-2-PiQ}&--&0.459&0.479&0.636&0.564&0.578&0.401&0.439&0.391&0.423&0.397\\
			\midrule
			\emph{CONTRIQUE-FR}&$\times$&0.889&0.898&0.855&0.843&0.666&0.645&0.732&0.730&0.469&0.484\\
			\midrule
			\emph{FPR}&--&0.826&0.895&0.793&0.735&0.576&0.519&0.530&0.528&0.313&0.316\\
			\midrule
			\emph{DISTS}&$\times$&0.945&0.948&0.923&0.920&0.755&0.708&--&--&0.617&0.596\\
			\midrule
			\emph{BIQA-M.D.}&--&0.834&0.876&0.762&0.678&0.479&0.420&--&--&--&--\\
			\midrule
			\emph{SSHMPQA}&$\times$&\underline{0.959}&\underline{0.963}&0.945&0.945&\textbf{0.897}&\textbf{0.879}&--&--&0.653&0.634\\
			\midrule
			DeepSSIM-Lite&$\checkmark$&0.910&0.955&0.935&0.939&0.829&0.797&\underline{0.875}&\underline{0.877}&\underline{0.710}&\underline{0.683}\\
			
			DeepSSIM&$\checkmark$&\underline{0.959}&\textbf{0.964}&\textbf{0.948}&\textbf{0.950}&\underline{0.860}&\underline{0.838}&\textbf{0.902}&\textbf{0.903}&\textbf{0.734}&\textbf{0.699}\\
			\bottomrule
	\end{tabular} }
\end{table*}   

\section{Experimental Results}  
\subsection{Experimental Setups}
\subsubsection{IQA Datasets}
We evaluate our method on eleven benchmark IQA datasets, including four AR-IQA datasets (LIVE \cite{LIVE}, CSIQ \cite{CSIQ}, TID2013 \cite{TID2013} and KADID-10k \cite{KADID}), six GDR-IQA datasets including three SRIQA datasets (CVIU \cite{CVIU}, QADS \cite{QADS} and SISAR \cite{SISAR}) and three IRQA datasets (CUHK \cite{CUHK}, RetargetMe \cite{RetargetMe} and NRID \cite{NRID}), and a challenging IQA dataset PIPAL \cite{PIPAL}. PIPAL is a large-scale IQA dataset for perceptual image restoration, which covers both AR-IQA and GDR-IQA tasks. Specifically, in addition to traditional distortion, it contains massive GAN-based algorithms distortion and denoising distortion. It should be noted that only the training set of the PIPAL dataset is publicly available. The details of the eleven datasets are shown in Table \ref{table:datasets}.
\subsubsection{Comparison Algorithms}
We compare our DeepSSIM metric with different algorithms that were designed for different IQA tasks. Notably, the public codes of majority IRQA methods are unavailable. Hence, towards non-training-based methods, we choose several typical algorithms and directly citing their published results. For training-based methods, we select comparisons that present cross-dataset test results since the generalization of IQA is crucial. To be specific, the comparison methods include ten general IQA methods (SSIM \cite{SSIM}, FSIM \cite{FSIM}, GMSD \cite{GMSD}, LPIPS \cite{LPIPS}, PaQ-2-PiQ \cite{PaQ-2-PiQ}, CONTRIQUE \cite{CONTRIQUE}, FPR \cite{FPR}, DISTS \cite{DISTS}, BIQA-M.D. \cite{BIQAMD} and SSHMPQA \cite{SSHMPQA}), six popular SRIQA methods (HYQM \cite{HYQM}, CNNSR \cite{CNNSR}, DeepSRQ \cite{DeepSRQ}, SFSN \cite{SFSN}, DISQ \cite{SISAR} and PSCT \cite{PSCT}) and seven typical IRQA algorithms (EMD \cite{EMD}, CSim \cite{CSim}, PGDIL \cite{NRID}, ARS \cite{ARS}, MLF \cite{MLF}, Jiang \cite{Jiang2020} and INSEM \cite{INSEM}). 

\subsubsection{Evaluation Criteria}
For the IQA datasets labeled with MOS or DMOS, we use two common criteria including PLCC and SRCC, which measure the prediction accuracy and monotonicity, respectively. For the IQA datasets labeled with the number of votes (RetargetMe and NRID), a prediction accuracy cannot be directly calculated. Following the related IRQA methods, we employ Kendall Rank Correlation Coefficient (KRCC) to measure the correlation between subjective rank and objective rank. In practice, we adopt the mean and standard deviation KRCC values computed from the different sets of retargeted images.
\begin{table}[t]
	\centering
	\setlength{\abovecaptionskip}{0cm}
	\setlength{\belowcaptionskip}{-0.2cm}
	\setlength{\tabcolsep}{1mm}{
	\caption{Performance comparison on SRIQA datasets. We mark the training-based
		methods in italics and present their generalization performance. The top two results are labeled in bold and underline}
	\label{table:sr}
	\begin{tabular}{@{}c|c|cc|cc|cc@{}}
		\toprule
		&& \multicolumn{2}{c|}{CVIU}&\multicolumn{2}{c|}{QADS}&\multicolumn{2}{c}{SISAR}   \\ 
		\cmidrule(l){3-8} 
		\multirow{-2}{*}{Algorithms}&\multirow{-2}{*}{Ref.} &PLCC & SRCC  & PLCC & SRCC	& PLCC & SRCC\\ 
		\midrule
		\emph{CNNSR}&$\times$&0.539&0.542&0.382&0.364&0.290&0.236\\
		\midrule
		\emph{DeepSRQ}&$\times$&0.670&0.650&0.564&0.516&0.545&0.460\\
		\midrule
		HYQM&LR&--&--&0.317&0.434&0.461&0.515\\
		\midrule
		\emph{DISQ}&LR&--&--&0.730&0.722&0.575&0.448\\
		\midrule
		SFSN&HR&0.839&0.825&0.802&0.800&0.640&0.638\\
		\midrule
		&LR&--&--&0.655&0.612&0.575&0.568\\
		\cmidrule(l){2-8}
		\multirow{-2}{*}{\emph{PSCT}}&HR&0.752&0.750&0.815&0.807&0.664&0.647\\
		\midrule
		&LR&--&--&0.599&0.589&0.660&0.653\\
		\multirow{-2}{*}{DeepSSIM-Lite}&HR&\underline{0.896}&\underline{0.871}&\underline{0.828}&\underline{0.821}&\underline{0.726}&\underline{0.711}\\
		\cmidrule(l){2-8}

		&LR&--&--&0.741&0.726&0.683&0.677\\
		\multirow{-2}{*}{DeepSSIM}&HR&\textbf{0.911}&\textbf{0.891}&\textbf{0.846}&\textbf{0.840}&\textbf{0.760}&\textbf{0.750}\\
		\bottomrule
	\end{tabular}}
\end{table}

\begin{table}[t]
	\centering
	\setlength{\abovecaptionskip}{0cm}
	\setlength{\belowcaptionskip}{-0.2cm}
	\setlength{\tabcolsep}{1mm}{
	
		\caption{Performance comparison on IRQA datasets. We mark the training-based methods in italics and present their generalization performance. The top two results are labeled in bold and underline}
		\label{table:re}
		\begin{tabular}{@{}c|cc|cc|cc@{}}
			\toprule
			& \multicolumn{2}{c|}{CUHK}&\multicolumn{2}{c|}{RetargetMe}&\multicolumn{2}{c}{NRID}  \\ 
			\cmidrule(l){2-7} 
			\multirow{-2}{*}{Algorithms}&PLCC & SRCC  &
			\makecell{Mean\\KRCC}	&\makecell{Std\\KRCC} &\makecell{Mean\\KRCC}
			&\makecell{Std\\KRCC}\\ 
			\midrule
			EMD&0.276&0.290&0.251&0.272&0.362&0.361\\
			\midrule
			CSim&0.437&0.466&0.164&0.263&--&--\\
			\midrule
			PGDIL&0.540&0.541&0.415&0.296&--&--\\
			\midrule
			ARS&\underline{0.684}&\underline{0.669}&0.452&0.283&0.514&0.398\\
			\midrule
			\emph{MLF}&\textbf{0.730}&\textbf{0.713}&0.469&0.256&--&--\\
			\midrule
			\emph{Jiang\cite{Jiang2020}}&0.463&0.247&--&--&--&--\\
			\midrule
			\emph{INSEM}&--&--&\textbf{0.537}&0.188&\textbf{0.640}&0.433\\
			\midrule
			DeepSSIM-Lite&0.636&0.600&0.279&\textbf{0.175}&0.429&\underline{0.281}\\
			
			DeepSSIM&0.664&0.641&\underline{0.471}&\underline{0.177}&\underline{0.619}&\textbf{0.270}\\
			\bottomrule
	\end{tabular} }
\end{table}

\subsubsection{Implementation Details}
We present the generalization performance of all training-based IQA methods. Specifically, we present their generalization results by executing their released pre-trained models on testing set. For LIVE, CSIQ and TID2013 datasets, we adopt the entire dataset as the testing set. For comparison experiments on SRIQA datasets, we follow the experimental setup of \cite{PSCT}. Towards the large-scale datasets KADID-10k and PIPAL, we randomly select 1000 images five times as the testing sets and average their results as the reported performance.  

We show the quantitative results of DeepSSIM metric and comparison methods on AR-IQA datasets and PIPAL dataset in Table \ref{table:traditional}. From the results, we have the following observations. First, the proposed DeepSSIM achieves the optimal performance on AR-IQA datasets. Second, only the proposed DeepSSIM metric can support input images with arbitrary resolution compared with other reference-based IQA methods. 

We show the quantitative results of DeepSSIM metric, SRIQA and IRQA comparison methods on GDR-IQA datasets in Table \ref{table:sr} and Table \ref{table:re}, respectively. From the results, we have the following observations. First, our DeepSSIM metric achieves the optimal performance on SRIQA datasets. Second, DeepSSIM metric reaches competitive performance on IRQA datasets. Third, the generalization performance of learning-based SRIQA and IRQA methods are unsatisfactory.

\section{Conclusions}
In this paper, we proposed a unified, non-training-based DeepSSIM index that deviates from existing approaches designed for specific geometrical distortion cases, but is capable of assessing image quality with and without geometric transformations, and geometric distortions of different types and at different levels. Extensive experimental results have demonstrated the superiority of our DeepSSIM metric, which can handle AR-IQA and GDR-IQA tasks simultaneously and outperforms the state-of-the-arts. Besides, our tests also reveal the effectiveness of DeepSSIM as an optimization tool for training computer vision tasks such as image super-resolution, enhancement and restoration. We envision a practical use of DeepSSIM metric in multi-task image processing systems.

\balance
\bibliographystyle{IEEEtran}
\bibliography{./ref1}%

\end{document}